# A LiDAR-Inertial SLAM Tightly-Coupled with Dropout-Tolerant GNSS Fusion for Autonomous Mine Service Vehicles

Yusheng Wang, *Graduate Student Member, IEEE*, Yidong Lou, Weiwei Song, Bing Zhan, Feihuang Xia and Qigeng Duan

*Abstract*—Multi-modal sensor integration has become a crucial prerequisite for the real-world navigation systems. Recent studies have reported successful deployment of such system in many fields. However, it is still challenging for navigation tasks in mine scenes due to satellite signal dropouts, degraded perception, and observation degeneracy. To solve this problem, we propose a LiDAR-inertial odometry method in this paper, utilizing both Kalman filter and graph optimization. The front-end consists of multiple parallel running LiDAR-inertial odometries, where the laser points, IMU, and wheel odometer information are tightly fused in an error-state Kalman filter. Instead of the commonly used feature points, we employ surface elements for registration. The back-end construct a pose graph and jointly optimize the pose estimation results from inertial, LiDAR odometry, and global navigation satellite system (GNSS). Since the vehicle has a long operation time inside the tunnel, the largely accumulated drift may be not fully by the GNSS measurements. We hereby leverage a loop closure based re-initialization process to achieve full alignment. In addition, the system robustness is improved through handling data loss, stream consistency, and estimation error. The experimental results show that our system has a good tolerance to the long-period degeneracy with the cooperation different LiDARs and surfel registration, achieving meter-level accuracy even for tens of minutes running during GNSS dropouts.

## I. Introduction

THE continuous spread of COVID-19 has promoted a growing demand of robotics in a great variety of scenes, from hospitals, construction sites, assembly plants, to mines. This surge of interest is motivated by a wide range of unmanned applications, such as autonomous service robots and robotaxis. The deployment of autonomous mine vehicles is of particular interest since the potential benefits include access to unreachable or dangerous locations and monitoring personnel in unsafe areas. These peculiarities will have affirmative impact on the mine operation, production, and safety.

Localization and environment perception are the essential capabilities for autonomous mine vehicles operation. In typical GPS-denied underground mine environments, many researches have proposed to use simultaneously localization and mapping (SLAM) to solve these problems [1], [2]. Unfortunately, most mature SLAM approaches have undesirable performance when deployed in real-life mine environments: the poor illumination renders visual-SLAM systems unreliable, the slippery terrain makes the wheel odometry inaccurate, and the explosion-proof requirements limit the large deployment of wireless sensors such as UWB and RFID.

As light detection and ranging (LiDAR) sensors are less sensitive to illumination variations and provide direct, high-fidelity, long-range 3D measurements, they have been widely accepted for odometry estimation in the past decade. Typically, LiDAR odometry algorithms estimate the ego motion of the vehicle through registering consecutive LiDAR frames. When it comes to perceptually-challenging mine environments, the presence of self-repetitive and symmetric areas increases the difficulty of constraining the relative motion along the main shaft of the mine tunnel. Techniques employed to mitigate this issue consist of observability analysis [3], degeneracy detection and mitigation [4], [5] and the integration of other sources of measurements, such as inertial data from an IMU.

The challenges of autonomous mine vehicles SLAM extend to engineering implementation. SLAM algorithms must operate onboard with limited computational budget, and deliver vehicle pose estimations with low latency regularly. Moreover, these SLAM systems are required to withstand intermittent sensor measurements and recover from transitory faulty states.

In this paper, we present a LiDAR odometry system that enables robust and accurate state estimation and environment reconstruction for autonomous mine vehicles. Concretely, the contributions of this paper include:

1) We develop a LiDAR-inertial system that incorporates the

*Manuscript submitted Dec 10, 2022. This work was supported by the Joint Foundation for Ministry of Education of China under Grant 6141A0211907 (*Corresponding author*: Yidong Lou).

Yusheng Wang is with the GNSS Research Center, Wuhan University, 129 Luoyu Road, Wuhan 430079, China and the CHC NAVIGATION, 6 Huanlongshandong Road, Building 2, 430073, Wuhan, China and the Beijing Lishedachuan Co., Ltd, 5 Yiheyuan Road, Beijing 100871, China (email: yushengwhu@whu.edu.cn).

Yidong Lou and Weiwei Song are with the GNSS Research Center, Wuhan University, 129 Luoyu Road, Wuhan 430079, China (email: ydlou@whu.edu.cn; sww@whu.edu.cn).

Bing Zhan is with the CHC NAVIGATION, 599 Gaojing Road, Building D, 201702, Shanghai, China. (email: zhanbing@huace.cn).

Feihuang Xia is with the Beijing Lishedachuan Co., Ltd, 5 Yiheyuan Road, Beijing 100871, China (email: xiafeihuang@dachuantech.net).

Qigeng Duan is with the Beijing Lishedachuan Co., Ltd, 5 Yiheyuan Road, Beijing 100871, China and the Department of Geography and Resource Management, The Chinese University of Hongkong, Hong Kong Special Administrative Region, Hong Kong 999077, China. (email: duanqigeng@gmail.com).

information from two LiDARs, an IMU, and wheel odometers using error-state Kalman Filter (ESKF) and graph optimization. Instead of using the commonly used iterated closet point (ICP) or feature points for registration, we merge laser scans through surfel fusion.
2) To fully compensate the largely accumulated drifts inside the tunnel, we develop a loop closure aided re-initialization method after long periods of GPS-dropouts. In that case, an estimation of the accumulated drift during the GPS outages is provided, then the errors both inside and outside the tunnel can be well eliminated.
3) The proposed pipeline is thoroughly evaluated in various mine tunnel environments across a long-time span. The results show our system drifts only 1.86 m after travelling up to 6.6 km (5 km tunnel).

The reminder of this paper is organized as follows. Section II reviews the relevant scholarly works. Section III gives an overview of the proposed system. Section IV presents the detailed graph optimization process applied in our system, followed by the experimental results described in Section V. Finally, Section VI concludes this paper and demonstrates future research directions.

## II. Related Work

Prior works on point cloud registration and LiDAR SLAM in tunnel-like environments are extensive. In this section, we briefly review scholarly works on these two aspects.

### A. Point Cloud Registration

The point cloud registration calculates the frame-to-frame displacement and matches the consecutive scans based thereon. It can be broadly classified into three different categories: point based, feature based and mathematical property based methods. The point based methods can be treated as dense approaches since they make full use of points from raw LiDAR scans. On the other hand, the feature-based approaches are regarded as sparse methods, as they only employ a select number of points for tracking. Furthermore, the mathematical property methods take advantage of statistical models, and transform the discrete representations of a single scan into a continuous distribution.

Many point based methods are the variations of ICP [6], which is an iterative two-step process. The algorithm first establishes the correspondences between the source and target point clouds. Then a transformation is calculated to reduce the distance between all corresponding points. Both step is repeated until reaching preset criteria. Considering its large computation burden with increased point cloud numbers, many approaches have tried to reduce its cost for real-time operation [7]–[9].

The feature based methods receive growing interest in recent years due to their simplicity and relatively lower computational complexity. Through extracting and matching feature points on planar surface and edges from the current and previous scan, the relative motion can be estimated accordingly. Similar to the Dilution of Precision (DOP) concept in the field of satellite navigation, the feature distribution also has a great influence to the state estimation results. The frame-to-frame registration is prone to fail once the environment is mostly planar. Therefore, some methods [10]–[12] use surfel as small planar feature and register new points by minimizing the point to plane residuals. In this paper, we add the degeneracy analysis [5] to the surfel matching to further improve the matching accuracy inside the tunnels, where the state estimation problem is only performed in the well-conditioned direction.

The normal distribution transform (NDT) [13] is a widely used mathematical property based methods. NDT divides the 3D space into small cells, and calculate the local probability density function (PDF) in each cell. Then the point-to-distribution correspondences are computed within a scan pair to find the optimal transformation. NDT aims to maximize the likelihood of all points described by the normal distributions. This reduces the memory consumption as well as computation time for nearest-neighbor searches.

### B. LiDAR SLAM in Tunnel-like Environments

The current LiDAR SLAM have proved to be accurate and robust enough for many scenarios, and we are mainly focused in handling degeneracy here. LiDAR SLAM usually produces a large drift error in scenes with textureless surface or repeated structures, such as indoor environments or tunnels. Researchers have proposed adding auxiliary sensors, degeneracy analysis, and geometric structures to cope with this problem.

Adding sensor means introducing additional constraints. Cameras have proved to be a good complementary sensor in some LiDAR degenerated districts [14]–[16], but they are still inaccurate in perceptually-degraded mine tunnels. As an environment-insensitive sensor, the ultra-wideband (UWB) provides less cumulative errors and has attracted more and more interest in recent years [17]. However, they need to be largely deployed along the mine tunnel, which does not satisfy the explosion-proof regulation of most mine tunnels. As discussed in our previous work [18], the LiDARs of limited field of view (FoV) but high point cloud density is less likely to fail at degenerated districts. On the other hand, the spinning LiDAR with 360° FoV can provide consistent state estimation towards irregular movement and better loop closing. In this paper, we leverage the advantages of both LiDARs in the system design.

One of the early works of degeneracy analysis is proposed by Zhang et al. [5], which leverages the minimum eigenvalue of the information matrices to determine system degeneracy. However, this metric is difficult to interpret because of unclear physical meaning. Zhen et al. define the localizability vector through projecting information matrix into the eigenspace, and model the degeneration with a frictionless force-closure [19].Tagliabue et al. [20] also use the smallest eigenvalue of point-to-plane cost to indicate the least observable direction. Instead of solving the degeneracy problem using the LiDAR-inertial SLAM directly, it will switch to other parallel-running odometry algorithms [21] when the metric is below a self-defined threshold. The degeneracy analysis based methods have been largely deployed to tunnel exploration tasks [20], [22]. We also introduce this degeneracy analysis feature into our system to further improve system accuracy.

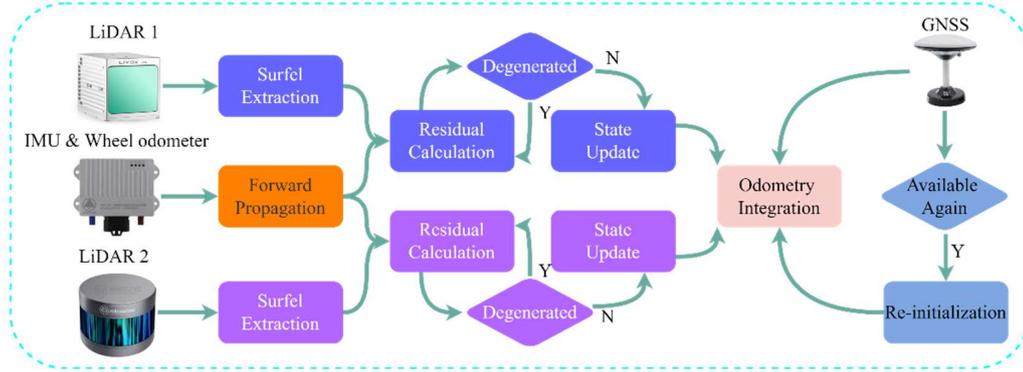

**Fig. 1.** The overview of the proposed system.

TABLE I
NOTATIONS THROUGHOUT THE PAPER

| Notations | Explanations |
|---|---|
| | Coordinates |
| $(\cdot)^W$ | The global frame used for vehicle navigation. |
| $(\cdot)^B$ | The body frame which is also the IMU frame. |
| $(\cdot)^O$ | The odometer frame, which is often expressed in B frame |
| $(\cdot)^L$ | The LiDAR frame, defined by the manufacturer. |
| | Expression |
| $\mathbf{p}$ | The position or translation vector. |
| $\mathbf{R}, \mathbf{q}$ | Two forms of rotation expression, $\mathbf{R} \in SO(3)$ is the rotation vector, $\mathbf{q}$ represents quaternions. |
| $\mathbf{v}$ | The linear velocity vector. |
| $\mathbf{x}$ | The vehicle full state vector. |
| $\mathbf{Z}$ | The full set of measurements. |
| $\mathbf{n}$ | The measurement noise. |

Man-made environments are often with strong structural regularity, lines, surfaces and objects. These geometric features have been widely exploited int LiDAR SLAM to improve state estimation accuracy [23]–[25]. Zhou et al. adopt the planar constraints to optimize plane parameters in the back-end [26], achieve promising results in a single-layer indoor environment. Zhou et al. [27] propose to use the principal component analysis (PCA) to extract sphere features along with planar, edge, and ground features. The results show that the spherical features can improve the system stability in highway scenarios.

## III. TIGHTLY COUPLED LIDAR-INERTIAL SLAM

The pipeline of the proposed system is visualized in Fig. 1. The two LiDARs are sent to different state estimation module, which estimates the full LiDAR state by registering surfels in a scan to the map via a tightly coupled ESKF. Then the weights are calculated accordingly and integrated with the GNSS measurements. Once the GNSS signal is available again after long dropouts, the re-initialization module is awakened to further optimize the trajectory and the mapping result.

Before diving into details of methods, we first define the frames and notations used throughout this article in TABLE I. In addition, we denote $(\cdot)^B_L$ as the transformation from LiDAR frame to IMU frame. Besides, we follow the "boxplus" and "boxminus", $\boxplus$ and $\boxminus$, operation defined in [28] to parameterize the state error on manifold. For a manifold $\mathcal{M}$ with dimension $n$, we have:

$$\mathcal{M} = SO(3): \quad \mathbf{R} \boxplus \mathbf{r} = \mathbf{R}\text{Exp}(\mathbf{r}); \mathbf{R}_1 \boxminus \mathbf{R}_2 = \text{Log}(\mathbf{R}_2^T \mathbf{R}_1)$$

$$\mathcal{M} = \mathbb{R}^n: \quad \mathbf{a} \boxplus \mathbf{b} = \mathbf{a} + \mathbf{b}; \quad \mathbf{a} \boxminus \mathbf{b} = \mathbf{a} - \mathbf{b}$$

$$\text{Exp}(\mathbf{r}) = \mathbf{I} + \frac{\mathbf{r}}{\|\mathbf{r}\|}\sin(\|\mathbf{r}\|) + \frac{\mathbf{r}^2}{\|\mathbf{r}\|^2}(1 - \cos(\|\mathbf{r}\|)), \quad (1)$$

where $\text{Exp}(\mathbf{r})$ is the exponential map in [28] and $\text{Log}(\cdot)$ is its inverse map. Therefore, we have the following expression for a compound manifold $\mathcal{M} = SO(3) \times \mathbb{R}^n$,

$$\begin{bmatrix}\mathbf{R}\\\mathbf{a}\end{bmatrix} \boxplus \begin{bmatrix}\mathbf{r}\\\mathbf{b}\end{bmatrix} = \begin{bmatrix}\mathbf{R} \boxplus \mathbf{r}\\\mathbf{a}+\mathbf{b}\end{bmatrix}; \begin{bmatrix}\mathbf{R}_1\\\mathbf{a}\end{bmatrix} \boxminus \begin{bmatrix}\mathbf{R}_2\\\mathbf{b}\end{bmatrix} = \begin{bmatrix}\mathbf{R}_1 \boxminus \mathbf{R}_2\\\mathbf{a}-\mathbf{b}\end{bmatrix}. \quad (2)$$

Utilizing the definitions above, we can easily derive:

$$\boxplus: \mathcal{M} \times \mathbb{R}^n \to \mathcal{M}; \boxminus: \mathcal{M} \times \mathcal{M} \to \mathbb{R}^n,$$

$$(\mathbf{x} \boxplus \mathbf{u}) \boxminus \mathbf{x} = \mathbf{u}; \mathbf{x} \boxplus (\mathbf{y} \boxminus \mathbf{x}) = \mathbf{y}; \mathbf{x}, \mathbf{y} \in \mathcal{M}, \mathbf{u} \in \mathbb{R}^n \quad (3)$$

Besides, for the vehicle full state $\mathbf{x}$, we denote the following definitions used in the iterated Kalman filter:

- $\mathbf{x}$    The ground true value of state $\mathbf{x}$.
- $\hat{\mathbf{x}}$    The propagated value of state $\mathbf{x}$.
- $\bar{\mathbf{x}}$    The updated value of state $\mathbf{x}$.
- $\tilde{\mathbf{x}}$    The error between the ground true $\mathbf{x}$ and its estimation $\hat{\mathbf{x}}$.
- $\hat{\mathbf{x}}^\kappa$    The estimation of state $\mathbf{x}$ in the $\kappa$-th iteration of the iterated Kalman filter.

### A. State Transition Model

The raw accelerometer and gyroscope measurements, $\hat{\mathbf{a}}$ and $\hat{\boldsymbol{\omega}}$, are given by:

$$\hat{\mathbf{a}}_k = \mathbf{a}_k + \mathbf{R}_W^{B_k}\mathbf{g}^W + \mathbf{b}_{a_k} + \mathbf{n}_a,$$
$$\hat{\boldsymbol{\omega}}_k = \boldsymbol{\omega}_k + \mathbf{b}_{\omega_k} + \mathbf{n}_\omega, \quad (4)$$

where $\mathbf{n}_a$ and $\mathbf{n}_\omega$ are the zero-mean white Gaussian noise, with $\mathbf{n}_a \sim \mathcal{N}(\mathbf{0}, \boldsymbol{\sigma}_a^2)$, $\mathbf{n}_\omega \sim \mathcal{N}(\mathbf{0}, \boldsymbol{\sigma}_\omega^2)$. The gravity vector in the world frame is denoted as $\mathbf{g}^W = [0,0,\text{g}]^T$.

The odometer mounted on the wheel is utilized to measure the longitudinal velocity of the train along the rails, and the model of odometer sensor is given by:

$$\mathbf{c}^{O_k}\hat{\mathbf{v}}^O = \mathbf{v}^O + \mathbf{n}_{s^O}, \quad (5)$$

where $\mathbf{c}^{O_k}$ denotes the scale factor of the odometer modeled as random walk, with $\mathbf{n}_{s^O} \sim \mathcal{N}(\mathbf{0}, \boldsymbol{\sigma}_{s^O}^2)$. Then the pose estimation can be achieved through synchronously collected gyroscope and odometer output, and the displacement within two consecutive frames $k$ and $k+1$ can be given as:

$$\hat{\mathbf{p}}_{O_k}^{O_{k+1}} = \mathbf{p}_{O_k}^{O_{k+1}} + \mathbf{n}_\mathbf{p}o, \tag{6}$$

where $\mathbf{n}_\mathbf{p}o$ is also the zero-mean white Gaussian noise. Based thereupon, we can derive the kinematic model as:

$$\dot{\mathbf{R}}_{B_k}^W = \mathbf{R}_{B_k}^W \lfloor \hat{\boldsymbol{\omega}}_k - \mathbf{b}_{\omega_k} - \mathbf{n}_\omega \rfloor_\wedge, \quad \dot{\mathbf{p}}_{B_k}^W = \mathbf{v}_{B_k}^W,$$
$$\dot{\mathbf{v}}_{B_k}^W = \mathbf{R}_{B_k}^W(\hat{\mathbf{a}}_k - \mathbf{b}_{a_k} - \mathbf{n}_a) + \mathbf{g}^W,$$
$$\dot{\mathbf{b}}_{\omega_k} = \mathbf{n}_{\mathbf{b}_\omega}, \dot{\mathbf{b}}_{a_k} = \mathbf{n}_{\mathbf{b}_a},$$
$$\mathbf{g}^W = \mathbf{0}, \dot{\mathbf{c}}^{O_k} = \mathbf{0}, \dot{\mathbf{R}}_{L_k}^{B_k} = \mathbf{0}, \dot{\mathbf{p}}_{L_k}^{B_k} = \mathbf{0}. \tag{5}$$

where $\mathbf{R}_{B_k}^W$, $\mathbf{p}_{B_k}^W$, and $\mathbf{v}_{B_k}^W$ denote the IMU attitude, position, and velocity expressed in the global frame. $\mathbf{b}_a$ and $\mathbf{b}_\omega$ are the IMU biases modeled as random walk process driven by $\mathbf{n}_{\mathbf{b}_a}$ and $\mathbf{n}_{\mathbf{b}_\omega}$. $\lfloor \kappa \rfloor_\wedge$ is the skew-symmetric cross product matrix of vector $\kappa \in \mathbb{R}^3$. The extrinsic between LiDAR and IMU is defined as $\mathbf{T}_{L_k}^{B_k} = \{\mathbf{R}_{L_k}^{B_k}, \mathbf{p}_{L_k}^{B_k}\}$. We can thereby derive the continuous model at the IMU sampling period $\Delta t$ can be discretized as:

$$\mathbf{x}_{i+1} = \mathbf{x}_i \boxplus \left(\Delta t \mathbf{f}(\mathbf{x}_i, \mathbf{u}_i, \mathbf{w}_i)\right), \tag{6}$$

where the state $\mathbf{x}_i$, input $\mathbf{u}_i$, process noise $\mathbf{w}_i$ and the function $\mathbf{f}$ are defined as:

$$\mathbf{x}_i =$$
$$[\mathbf{q}_{B_i}^W \quad \mathbf{p}_{B_i}^W \quad \mathbf{v}_{B_i}^W \quad \mathbf{b}_{g_i} \quad \mathbf{b}_{a_i} \quad \mathbf{g}^{W_i} \quad \mathbf{q}_{L_i}^B \quad \mathbf{p}_{L_i}^B \quad \mathbf{p}_{O_i}^B \quad \mathbf{c}^{O_i}]^T,$$
$$\mathbf{u}_i = [\hat{\boldsymbol{\omega}}_i \quad \hat{\mathbf{a}}_i \quad \hat{\mathbf{v}}^O{}_i]^T,$$
$$\mathbf{w}_i = [\mathbf{n}_{\omega_i} \quad \mathbf{n}_{a_i} \quad \mathbf{n}_{\mathbf{b}_{\omega_i}} \quad \mathbf{n}_{\mathbf{b}_{a_i}}]^T,$$

$$\mathbf{f}(\mathbf{x}, \mathbf{u}, \mathbf{w}) = \begin{bmatrix} \hat{\boldsymbol{\omega}} - \mathbf{b}_\omega - \mathbf{n}_\omega \\ \mathbf{v}_B^W + \frac{1}{2}(\mathbf{R}_B^W(\hat{\mathbf{a}} - \mathbf{b}_a - \mathbf{n}_a) + \mathbf{g}^W)\Delta t \\ \mathbf{R}_B^W(\hat{\mathbf{a}} - \mathbf{b}_a - \mathbf{n}_a) + \mathbf{g}^W \\ \mathbf{n}_{\mathbf{b}_\omega} \\ \mathbf{n}_{\mathbf{b}_a} \\ \mathbf{0}_{3\times 1} \\ \mathbf{0}_{3\times 1} \\ \mathbf{0}_{3\times 1} \\ \mathbf{R}_O^B(\mathbf{c}^{O_k}\hat{\mathbf{v}}^O - \mathbf{n}_s o) \\ \mathbf{0}_{3\times 1} \end{bmatrix}. \tag{7}$$

*B. Point Cloud Processing*

Two LiDARs, one Livox Avia and one Robosense RS-16 LiDAR, are included in our system. The former one is a hybrid solid-state LiDAR with non-repetitive scan pattern, while the latter one is a common mechanical spinning LiDAR.

For each LiDAR, we extract surfels through clustering points based on their positions and timestamps, then fitting ellipsoids to them. We first divide the 3D space into voxels and cluster LiDAR points within each voxel and with adjacent timestamps together. Secondly, we fit an ellipsoid to each sufficiently large point sets based on an empirical threshold. These ellipsoids are surfels extracted, with their centers and shapes determined by the sample mean and covariance of points in the cluster.

We employ a multi-resolution approach where the clustering and surfel extraction processes are repeated for multiple voxel sizes. More explicitly, we first cut the space into voxels, each with the size of the preset coarse map resolution. Thus, for each LiDAR scan, the contained points are distributed into voxels and are indexed into a Hash table. If all the contained points within a voxel lie on a plane, we store the plane points and calculate the surfel parameters. Each surfel is represented by its normal vector $\mathbf{n}$, which is the eigenvector corresponding to the smallest eigenvalue of the covariance matrix, and the center position $\mathfrak{p}$. Otherwise, the current voxel will break into eight octants and repeat plane checking until reaching the threshold. This process makes the voxels having different sizes, each voxel contains one plane feature fitted from the raw LiDAR points. To reduce the computational time cost, we measure the planarity of surfel by computing a score based on the spectrum of its covariance matrix and merely retain surfels that are sufficiently planar. Given a LiDAR point $\mathbf{p}_i^W$ accumulated in the world frame, we first search its nearest surfel. Then the point-to-plane distance can be calculated by:

$$d_i = \mathfrak{n}_i^T(\mathbf{p}_i^W - \mathfrak{p}_i). \tag{8}$$

In practice, each point $\mathbf{p}_i^L$ in the LiDAR scan contains both the ranging and beam-directing noises, denoted as $\mathbf{n}_i^L$. The real position of the point $\mathbf{p}_{gt_i}^L$ satisfies:

$$\mathbf{p}_{gt_i}^L = \mathbf{p}_i^L + \mathbf{n}_i^L. \tag{9}$$

Considering the corresponding IMU state expressed in the global frame, $\mathbf{T}_{B_i}^W = \{\mathbf{R}_{B_i}^W, \mathbf{p}_{B_i}^W\}$, and LiDAR-IMU extrinsic $\mathbf{T}_{L_i}^{B_i}$, we can obtain the true position of $\mathbf{p}_i^W$ through:

$$\mathbf{p}_i^W = \mathbf{T}_{B_i}^W \mathbf{T}_{L_i}^{B_i}(\mathbf{p}_i^L + \mathbf{n}_i^L). \tag{10}$$

Submitting (10) into (8), we can get the measurement model:

$$\mathbf{h}_i(\mathbf{x}_i, \mathbf{n}_i^L) = \mathfrak{n}_i^T \left(\mathbf{T}_{B_i}^W \mathbf{T}_{L_i}^{B_i}(\mathbf{p}_i^L + \mathbf{n}_i^L) - \mathfrak{p}_i\right). \tag{11}$$

The estimated pose is used to register the new points to the global map. When the new points belong to an unpopulated voxel, it will construct the voxel. Otherwise, it will be added to an existing voxel and update the parameters within the voxel. An example of the surfel global map is visualized in Fig. 2.

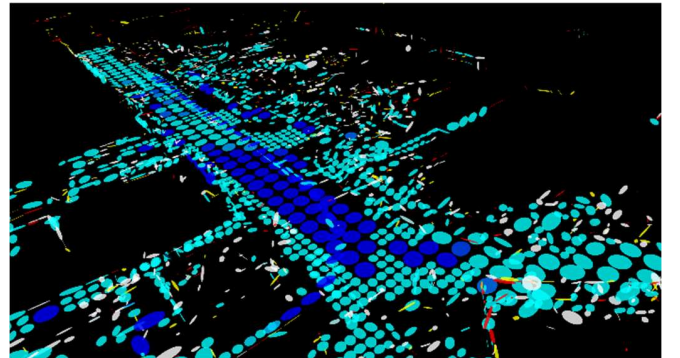

**Fig. 2.** An example of accumulated surfel map, where each surfel is represented by markers of different colors.

$$\mathbf{F}_{\tilde{\mathbf{x}}} = \begin{bmatrix} \mathrm{Exp}(-\widehat{\boldsymbol{\omega}}_i \Delta t) & 0 & 0 & -\mathbf{A}(\widehat{\boldsymbol{\omega}}_i \Delta t)^T \Delta t & 0 & 0 & 0 & 0 & 0 & 0 \\ 0 & \mathbf{I} & \mathbf{I}\Delta t & 0 & 0 & 0 & 0 & 0 & 0 & 0 \\ -\mathbf{R}_B^W \lfloor \widehat{\mathbf{a}}_i \rfloor_\wedge \Delta t & 0 & \mathbf{I} & 0 & -\mathbf{R}_B^W \Delta t & \mathbf{I}\Delta t & 0 & 0 & 0 & 0 \\ 0 & 0 & 0 & \mathbf{I} & 0 & 0 & 0 & 0 & 0 & 0 \\ 0 & 0 & 0 & 0 & \mathbf{I} & 0 & 0 & 0 & 0 & 0 \\ 0 & 0 & 0 & 0 & 0 & \mathbf{I} & 0 & 0 & 0 & 0 \\ 0 & 0 & 0 & 0 & 0 & 0 & \mathbf{I} & 0 & 0 & 0 \\ 0 & 0 & 0 & 0 & 0 & 0 & 0 & \mathbf{I} & 0 & 0 \\ 0 & 0 & 0 & 0 & 0 & 0 & 0 & 0 & \mathbf{I} & 0 \\ 0 & 0 & 0 & 0 & 0 & 0 & 0 & 0 & 0 & \mathbf{I} \end{bmatrix}, \mathbf{F}_\mathbf{w} = \begin{bmatrix} -\mathbf{A}(\widehat{\boldsymbol{\omega}}_i \Delta t)^T \Delta t & 0 & 0 & 0 \\ 0 & 0 & 0 & 0 \\ 0 & -\mathbf{R}_B^W \Delta t & 0 & 0 \\ 0 & 0 & \mathbf{I}\Delta t & 0 \\ 0 & 0 & 0 & \mathbf{I}\Delta t \\ 0 & 0 & 0 & 0 \\ 0 & 0 & 0 & 0 \\ 0 & 0 & 0 & 0 \\ 0 & 0 & 0 & 0 \\ 0 & 0 & 0 & 0 \end{bmatrix} \quad (13)$$

## C. Iterated Kalman Filter

For each LiDAR input, we employ an iterated Kalman filter to estimate the system state utilizing the state transition model (6) and measurement model (11). By setting the process noise to zero, we can perform the forward propagation upon received IMU data using the error state dynamic model [29]:

$$\widehat{\mathbf{x}}_{i+1} = \widehat{\mathbf{x}}_i \boxplus \left(\Delta t \mathbf{f}(\widehat{\mathbf{x}}_i, \mathbf{u}_i, \mathbf{0})\right)$$
$$= \mathbf{F}_{\tilde{\mathbf{x}}} \widetilde{\mathbf{x}}_i + \mathbf{F}_\mathbf{w} \mathbf{w}_i. \quad (12)$$

Here $\widehat{\mathbf{x}}_0 = \overline{\mathbf{x}}_{k-1}$, where $\overline{\mathbf{x}}_i$ is the optimal state estimation of the LiDAR scan at $t_i$. The matrix $\mathbf{F}_{\tilde{\mathbf{x}}}$ and $\mathbf{F}_\mathbf{w}$ is listed at the top of this page in (13). $\mathbf{A}(\mathbf{u})^{-1}$ follows the definition in [30]:

$$\mathbf{A}(\mathbf{u})^{-1} = \mathbf{I} - \frac{1}{2}\lfloor \mathbf{u} \rfloor_\wedge + (1 - \propto (\|\mathbf{u}\|)) \frac{\lfloor \mathbf{u} \rfloor_\wedge^2}{\|\mathbf{u}\|^2},$$
$$\propto (u) = \frac{u}{2}\frac{\cos(u/2)}{\sin(u/2)}. \quad (14)$$

Besides, the propagated covariance $\widehat{\mathbf{P}}_i$ can be calculated by:

$$\widehat{\mathbf{P}}_{i+1} = \mathbf{F}_{\tilde{\mathbf{x}}} \widehat{\mathbf{P}}_i \mathbf{F}_{\tilde{\mathbf{x}}}^T + \mathbf{F}_\mathbf{w} \mathbf{Q} \mathbf{F}_\mathbf{w}^T; \widehat{\mathbf{P}}_0 = \overline{\mathbf{P}}_{k-1}, \quad (15)$$

where $\mathbf{Q}_i$ is the covariance of the noise $\mathbf{w}_i$.

We need to compensate for the relative point cloud motion before they are integrated with the propagated state $\widehat{\mathbf{x}}_i$ and covariance $\widehat{\mathbf{P}}_i$ to produce an optimal state update. This process is implemented from [31], where (6) is propagated backward as

$$\check{\mathbf{x}}_{i-1} = \check{\mathbf{x}}_i \boxplus \left(-\Delta t \mathbf{f}(\mathbf{x}_i, \mathbf{u}_i, \mathbf{0})\right). \quad (16)$$

The backward propagation utilizes the left IMU and wheel odometer measurement as the input to compute a relative pose between point sampling time and the scan end time. After the motion compensation process, we can view all the points within the scan are sampled at the same time. Then we can derive the residual at the $\kappa$-th iteration $\mathbf{z}_i^\kappa$ as:

$$\mathbf{z}_i^\kappa = \mathbf{h}_i(\widehat{\mathbf{x}}_i^\kappa, \mathbf{0}) = \mathfrak{n}_i^T \left(\mathbf{T}_{B_i}^W \mathbf{T}_{L_i}^{B_i} \mathbf{p}_i^L - \mathfrak{p}_i\right). \quad (17)$$

The total measurement noise is $\mathfrak{n}_i^T \mathbf{T}_{B_i}^W \mathbf{T}_{L_i}^{B_i} \mathbf{n}_i^L \sim \mathcal{N}(\mathbf{0}, \mathbf{R}_i)$. Combining the prior distribution from the forward propagation $\mathbf{x}_i \boxminus \widehat{\mathbf{x}}_i$ and the measurement model, we can derive a posterior distribution of the state $\mathbf{x}_i$, denoted as $\widetilde{\mathbf{x}}_i^\kappa$, and its maximum a posterior (MAP) form:

$$\min_{\widetilde{\mathbf{x}}_i^\kappa} \left( \|\mathbf{x}_i \boxminus \widehat{\mathbf{x}}_i\|_{\widehat{\mathbf{P}}_i}^2 + \sum_{j=1}^m \|d_j - \mathbf{H}_j \cdot (\mathbf{x}_i \boxminus \overline{\mathbf{x}}_i)\|_{\mathbf{R}_j}^2 \right). \quad (18)$$

Let $\mathbf{H} = \left[\mathbf{H}_1^{\kappa T}, \mathbf{H}_2^{\kappa T}, \ldots, \mathbf{H}_m^{\kappa T}\right]^T$, $\mathbf{R} = \mathrm{diag}(\mathbf{R}_1, \mathbf{R}_2, \ldots, \mathbf{R}_m)$, $\mathbf{P} = (\mathbf{J}^\kappa)^{-1} \widehat{\mathbf{P}}_i (\mathbf{J}^\kappa)^{-T}$, and $\mathbf{z}_i^\kappa = \left[\mathbf{z}_1^{\kappa T}, \mathbf{z}_2^{\kappa T}, \ldots, \mathbf{z}_m^{\kappa T}\right]^T$, this MAP problem can be solved by iterated Kalman filter:

$$\mathbf{K} = (\mathbf{H}^T \mathbf{R}^{-1} \mathbf{H} + \mathbf{P}^{-1})^{-1} \mathbf{H}^T \mathbf{R}^{-1},$$
$$\widehat{\mathbf{x}}_i^{\kappa+1} = \widehat{\mathbf{x}}_i^\kappa \boxplus \left(-\mathbf{K}\mathbf{z}_i^\kappa - (\mathbf{I} - \mathbf{K}\mathbf{H})(\mathbf{J}^\kappa)^{-1}(\widehat{\mathbf{x}}_i^\kappa \boxminus \widehat{\mathbf{x}}_i)\right), \quad (19)$$

where $\mathbf{K}$ is the Kalman gain, $\mathbf{H}$ is the Jacobin matrix of the measurement model $\mathbf{h}_i(\mathbf{x}_i, \mathbf{n}_i^L)$, and $\mathbf{J}^\kappa$ is the partial differentiation of $(\widehat{\mathbf{x}}_i^\kappa \boxplus \widetilde{\mathbf{x}}_i^\kappa) \boxminus \widehat{\mathbf{x}}_i$ w.r.t. $\widetilde{\mathbf{x}}_i^\kappa$ evaluated at zero.

This process repeats until convergence, $\|\widehat{\mathbf{x}}_i^{\kappa+1} \boxminus \widehat{\mathbf{x}}_i^\kappa\| < \epsilon$, then the optimal state and covariance estimation are:

$$\overline{\mathbf{x}}_i = \widehat{\mathbf{x}}_i^{\kappa+1}, \overline{\mathbf{P}}_i = (\mathbf{I} - \mathbf{K}\mathbf{H})\mathbf{P}. \quad (20)$$

The state update is then used to transform each scan point to the global frame and inserted into the map.

## D. Graph Optimization

We construct a pose graph at the back-end to integrate pose information from two LiDAR odometries, inertial odometry, GNSS, and detected loop closures. This state estimation process can be formulated as a maximum-a-posterior (MAP) problem. Given the measurements $\mathbf{z}_k$ and the history of states $\mathbf{x}_k$, the MAP problem can be formulated as:

$$\mathbf{x}_k^* = \underset{\chi_k}{\mathrm{argmax}}\, p(x_k|\mathbf{x}_k) \propto p(\mathbf{x}_0)p((\mathbf{z}_k|\mathbf{x}_k)) \quad (21)$$

If the measurements are conditionally independent, then (21) can be solved through least squares minimization:

$$\chi^* = \underset{\chi_k}{\mathrm{argmin}} \sum \sum_{i=1}^k \|r_i\|^2 \quad (22)$$

where $r_i$ is the residual of the error between the predicted and measured value.

For the sake of decreasing system memory usage and increasing computation efficiency, we employ the sliding window to keep a relative steady number of nodes in the local graph. Given a sliding window containing $k$ keyframes, $X = [x_1^T, x_2^T, \ldots, x_k^T]^T$, we maximize the likelihood of the measurements, and the optimal states can be acquired through solving the MAP problem:

$$\min_{\mathbf{x}}\{\|\boldsymbol{r}_\mathcal{P}\|^2 + \mathcal{W}_{inertial}\sum_{i=1}^{N_\mathcal{J}}\|\boldsymbol{r}_{\mathcal{J}_i}\|^2 + \mathcal{W}_{avia}\sum_{i=1}^{N_\mathcal{L}^{avia}}\boldsymbol{r}_{\mathcal{L}_i}^{avia}$$
$$+ \mathcal{W}_{rs}\sum_{i=1}^{N_\mathcal{L}^{rs}}\boldsymbol{r}_{\mathcal{L}_i}^{rs} + \mathcal{W}_{GNSS}\sum_{i=1}^{N_\mathcal{G}}\|\boldsymbol{r}_{\mathcal{G}_i}\|^2\} \quad (23)$$

where $\boldsymbol{r}_\mathcal{P}$ is the prior factor marginalized by Schur-complement [32], $\boldsymbol{r}_{\mathcal{J}_i}$ is the residual of IMU-odometer preintegration result [18]. $\boldsymbol{r}_{\mathcal{L}_i}^{avia}$ and $\boldsymbol{r}_{\mathcal{L}_i}^{rs}$ define the residual of Avia and RS-16 LiDAR odometry. Finally, the GNSS constraints is denoted by $\boldsymbol{r}_{\mathcal{G}_i}$. Note that we use manually established value for the LiDAR odometry covariance and directly use the GNSS covariance predict from the raw measurements. Since the residuals are expressed in different frames, we unify their expression in the inertial frame using the calibrated sensor extrinsic such that:

$$\boldsymbol{r}_{\mathcal{G}_i} = \mathbf{R}_W^B(\mathbf{p}^{W_i} - \mathbf{p}^{W_{i-1}} - \mathbf{p}_W^B), \quad (24)$$

where the extrinsic $\mathbf{T}_B^W = \{\mathbf{R}_W^B, \mathbf{p}_W^B\}$ transform the GNSS pose into inertial coordinates.

We denote $N_\mathcal{J}$, $N_\mathcal{L}^{avia}$, $N_\mathcal{L}^{rs}$, and $N_\mathcal{G}$ as the number of four factors within the sliding window. $\mathcal{W}$ with footnotes defines the respective weighting factors.

We assume the short-term inertial preintegration result is accurate and set it as the reference to calculate other weighting factors. For the short period k to k+1, we denote $\mathbf{p}_{\mathcal{J}_k}^{\mathcal{J}_{k+1}}$ and $\mathbf{p}_{\mathcal{L}_k}^{\mathcal{L}_{k+1}}$ as the pose estimation results of inertial and LiDAR odometry. Then the LiDAR odometry weighting factor can be expressed using:

$$\mathcal{W}_\mathcal{L} = \left(1 - \left(\frac{\|\mathbf{p}_{\mathcal{L}_k}^{\mathcal{L}_{k+1}}\| - \|\mathbf{p}_{\mathcal{J}_k}^{\mathcal{J}_{k+1}}\|}{\|\mathbf{p}_{\mathcal{J}_k}^{\mathcal{J}_{k+1}}\|}\right)^2\right)\mathcal{W}_{inertial} \quad (25)$$

The $\mathcal{W}_{GNSS}$, on the other hand, is a combination of the DOP value, satellite number, and the real-time kinematic (RTK) solution status.

*E. Re-initialization*

The autonomous mine service vehicle starts in the open district, then enters the mine tunnel and travels for a long time, finally leaves the tunnel. In that case, the vehicle state may have accumulated a significant amount of drift inside the tunnel. If we directly fuse the large error LiDAR odometry and GNSS measurements, the accumulated drift may be not fully or over compensated, leading to extra pose estimation errors. Therefore, we propose to use a two-step re-initialization process for drift elimination.

1) *Loop detection*: Once the vehicle leaves the tunnel, the GNSS signal is available again. This will awake a iris loop detection thread based on [33], with the global pose and mapping updated upon detected loop. Note that we merely employ the 360° spinning LiDAR for loop detection.
2) *Full recovery*: After the loop is founded and the RTK is at the fixed solution, we will add the global measurements to

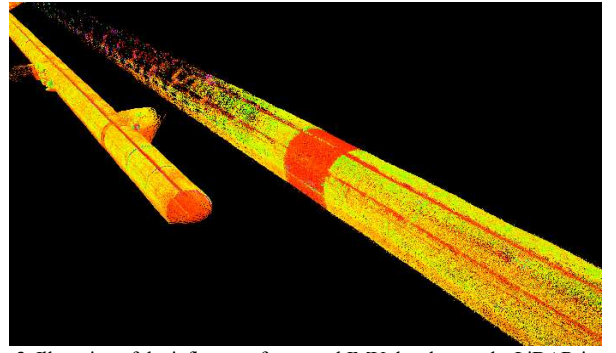

**Fig. 3.** Illustation of the influence of temporal IMU data loss to the LiDAR-inertial system. The real-time mapping has a sudden vertical drift w.r.t. a 0.6 s IMU data loss.

the pose graph as presented in Section IV-D. After this full recovery, the estimated state is again aligned with global.

*F. Hardware and Software Level Verification*

The hardware-level verification is conducted at the data preprocessing stage, including data stream existence, frequency, and individual verification. The data stream existence test aims to find out whether the required data input exist or not. Since the LiDAR-inertial odometry has a filter-based structure, and it will fail immediately when no input is from either IMU or LiDAR. Therefore, each of our LiDAR-inertial odometry will reinitialize and restart when either IMU or LiDAR stream is lost for one second. Otherwise, the filter-based system may generate large and unrecoverable drift as visualized in Fig. 3. The data frequency test also follows this idea. The system set the stream with the lowest frequency as the primary input, and monitor the counts of other data within two consecutive frames continuously, e.g., the LiDAR is set as the primary input (10 Hz), and approximately twenty frames of IMU input (200 Hz) should be found within two successive LiDAR scans. Once this criterion is not hold for thirty seconds, the system will send a warning to the user interface for a manual check, e.g., a yellow warning sign on the central control screen.

The individual verification is mainly for the LiDAR sensors. Since the mine service vehicle operates in the narrow tunnel, once the vehicle is facing direct to the wall, the LiDAR-inertial odometry may fail against the textureless and flat terrain. Therefore, we monitor the Euclidian distance of the point clouds within each scan, if 70 % of the points are below two meters to the LiDAR, the current frame is discarded for pose estimation. Once this stage retains for more than ten seconds, the system will switch to inertial odometry temporarily.

The software-level test is performed for parallel pose estimation modules to remove clearly wrong results. We set the maximum speed of the vehicle as 30 km/h, and verify whether the displacement of each odometry is beyond this limit or not, e.g., once the displacements of two successive LiDAR odometry (10 Hz) is beyond 2.0 m, it will be discarded for pose graph construction, since it is clearly wrong pose estimation results. Similarly, we use the steering angle information to monitor the individual yaw estimation results.

## IV. Experiments

To evaluate the performance of the proposed method, we conducted experiments in Madiliang mine of Ordos, China. As visualized in Fig. 4(a), the mine service vehicle transport food, staff worker, and some necessities between ground office and underground mine face. The underground mine tunnel is around 2.5 km long, with many branches along the path to the mine face as pictured in Fig. 4(b). We collected the dataset utilizing several autonomous mine service vehicles as shown in Fig 4(c). The vehicle is composed of one Robosense RS-16 spinning LiDAR with 360°×30° FoV and one Livox Avia non-repetitive scanning LiDAR with 70.4°×77.2° FoV. We use a ASENSING INS570D integrated navigation unit to provide GNSS and inertial measurements. We also collect the wheel encoder readings of the two rear wheels. The localization ground truth is kept by a MAPSTNAV POS620 high precision integrated navigation system with fiber optic gyros. POS620 supports precise post processing procedure, which can achieve centimeter-level localization accuracy in the long tunnels. In addition, we manually setup several check points inside the tunnel (on the wall or on the floor) using total station to further verify the localization and mapping accuracy.

All our algorithms are implemented in C++ and executed in Ubuntu Linux using the ROS [34]. We use an onboard computer with two NVIDIA Jetson Xaiver NX for real-time processing in the vehicle. Since all our sensors are hardware synchronized, we record the GNSS timestamp for each SLAM pose output. Then we can directly evaluate the localization accuracy through timestamp matching.

*A. Ground Tests*

The first experiments seek to evaluate the positioning and mapping accuracy of our system in the open sky environment. We first stay still waiting for the RTK initialization, and our SLAM algorithm will automatically align the estimation coordinates with global coordinates in this stage. Since there are no open source algorithms which integrate two LiDARs of different mechanism currently, we select three state-of-the-art (SOTA) open source algorithms, Lio-sam [35], Lili-om [36], and Fast-lio2 [4], that apply to both spinning and non-repetitive LiDARs for comparison. For the latter two approaches, we add the GNSS constraints of our approach to the back-end utilizing GTSAM [37]. In addition, we record the same GNSS timestamp to the selected approaches for positioning evaluation.

The first sequence is in the open sky, 2.7 km in length with time duration of 417.5 s. We plot the mapping result of Livox Avia LiDAR in Fig. 5, in which the consistent and clear building edges indicating our method is of high precision globally. Besides, we also provide four close views in Fig. 5 for the readers to inspect the local registration accuracy. We denote INS570D as the direct output of the INS570D integrated navigation unit. The 3D root mean square error (RMS) and maximum error (MAX) is computed and reported in TABLE II. Since the RTK status is always at fixed point solution along the journey, all the methods can be well-constrained by the global position measurements, and achieve a comparable accuracy with the post processed results. The LiDAR SLAM approaches cannot achieve a better accuracy than the INS570D as the range measuring error of LiDAR sensor is above 2 cm. In addition, we plot the sequential position and attitude error curves of our approach in Figure 6. We can infer that the largest error is in the vehicle moving direction (x direction), which doubles that of the y and z directional errors. Therefore, we have reason to believe that the time synchronization within different sensors is not accurate enough. Besides, we find that the asynchronous communication and delay between ROS nodes influence the performance of the SLAM algorithms. Different computers will output non identical results, and the odometry result for the same data input is not completely identical for different trials.

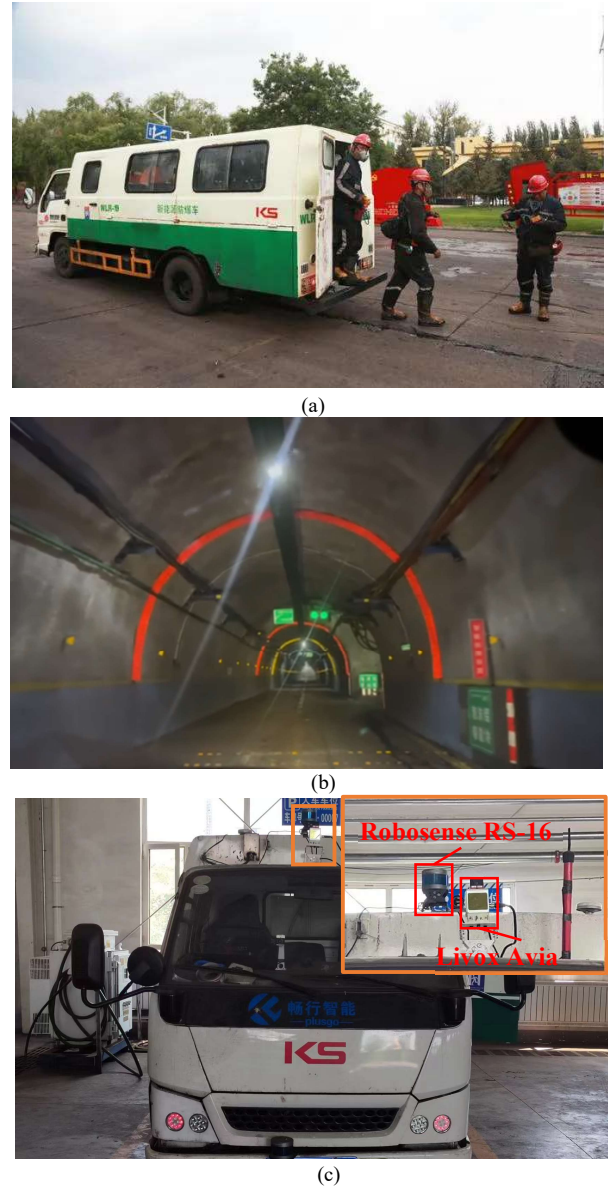

**Fig. 4**. Visualization of the mine service vehicle in (a) and an example of inner view of the mine tunnel.

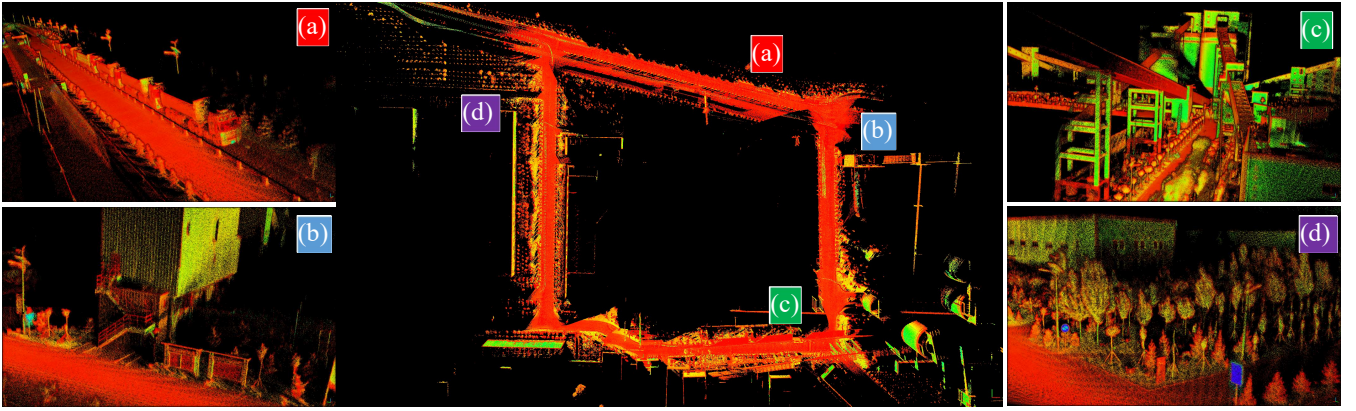

**Fig. 5**. The Livox Avia mapping result of ground test, sequence one. The top view of the overall mapping is pictured in the middle, (a), (b), (c), and (d) visualize the mapping in detail. (a) shows the mine trucks, (b) is the dormitory for miners, (c) presents the mine conveyor belt, and (d) is a crossing.

TABLE II
THE 3D LOCALIZATION ERROR W.R.T. GROUND TRUTH OF DIFFERENT APPROACHES.

| | RMSE [m] / MAX [m], with - and bold number indicates meaningless and best result, respectively. | | | | | | | |
|---|---|---|---|---|---|---|---|---|
| | Robosense RS-16 LiDAR | | | Livox Avia LiDAR | | | | |
| | Lio-sam | Lili-om | Fast-lio2 | Lio-sam | Lili-om | Fast-lio2 | INS570D | Ours |
| 1st sequence | 0.023/0.050 | 0.022/0.047 | 0.019/0.041 | 0.020/0.033 | 0.024/0.059 | 0.021/0.035 | **0.017/0.048** | 0.021/0.032 |
| 2nd sequence | 0.035/0.060 | 0.027/0.047 | 0.021/0.049 | 0.021/0.043 | 0.022/0.045 | 0.022/0.037 | 0.034/0.056 | **0.021/0.034** |

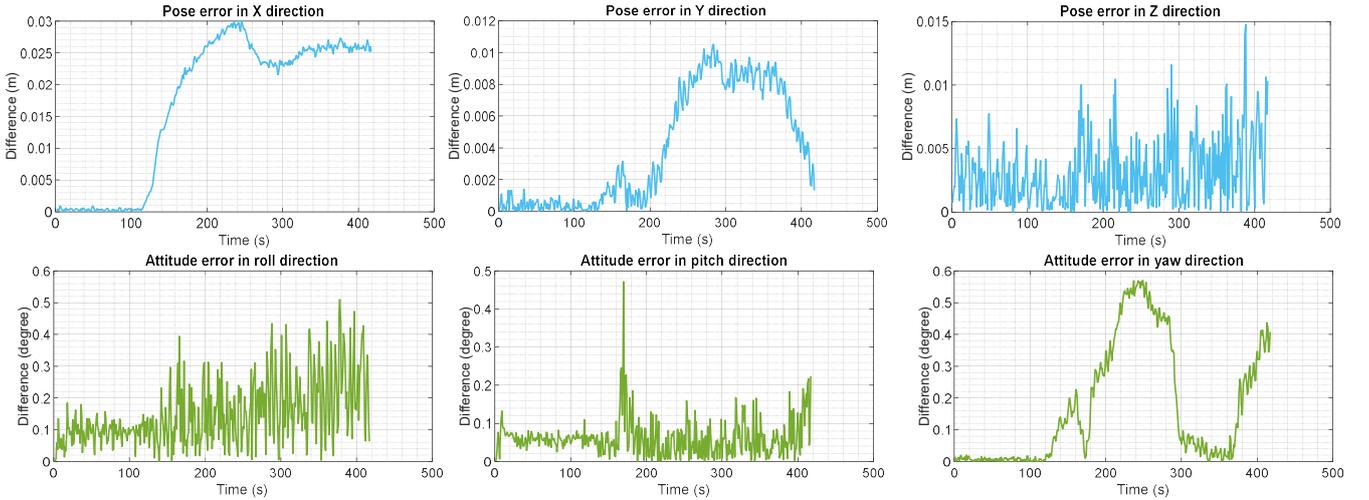

**Fig. 6**. The 3D positioning and 3D attitude error curves of our method for ground test, sequence one.

The second sequence is half indoor and half outdoor, 320 m in length with time duration of 278 s. As pictured in Fig. 7, the vehicle starts in the outdoor and slowly drives into the mining truck maintenance garage. The overall mapping result is visualized in Fig. 8, illustrating that the cooperate mapping benefits both the 360° coverage of Robosense RS-16 LiDAR and the high density of Livox Avia LiDAR. To further verify our mapping accuracy, we transform the local coordinates into WGS-84 and project the map onto satellite image as visualized in Fig. 9. The trees and building edges are well-matched to the background, demonstrating that our method is of high precision globally.

To quantitively present the localization accuracy of various methods, we compute the RMS and MAX errors and report them in TABLE II. The advantage of adding LiDAR sensor is now obvious, most of the approaches have an improvement of 25% than that of INS570D. This significant increment happens mainly indoors, where the GPS signal is blocked, and the IMU mechanization cannot hold for one minute. LiDAR SLAM now acts as a strong pose constraint at the GNSS outages, which significantly improve the localization accuracy.

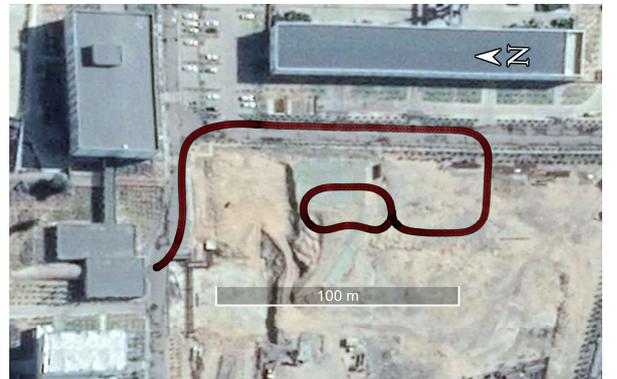

**Fig. 7**. The trajectory of our approach plotted onto satellite image.

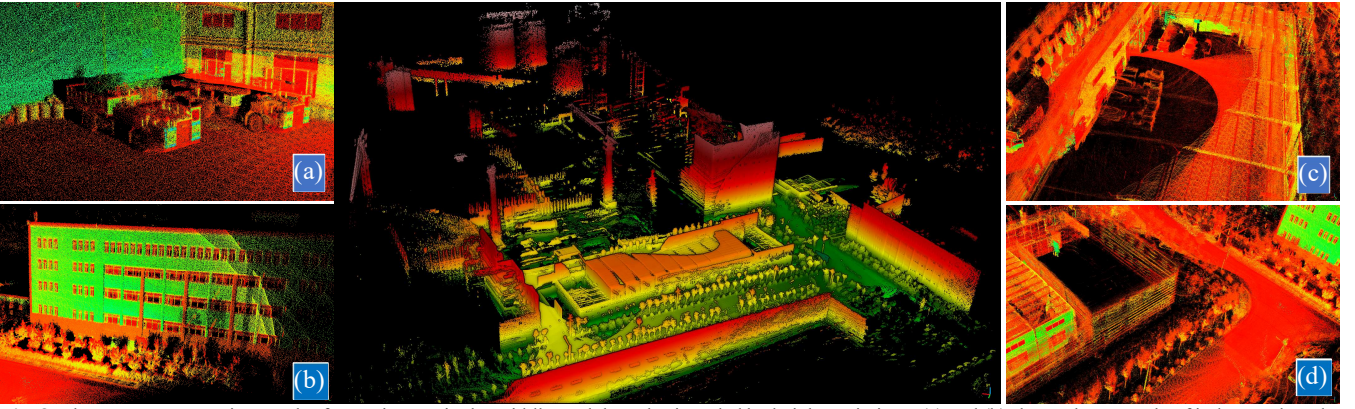

**Fig. 8**. The cooperate mapping result of two LiDARs in the middle, and the color is coded by height variations. (a) and (b) shows the example of indoor and outdoor mapping. (c) and (d) presents the advantage of the cooperate mapping, where both coverage and density is ensured.

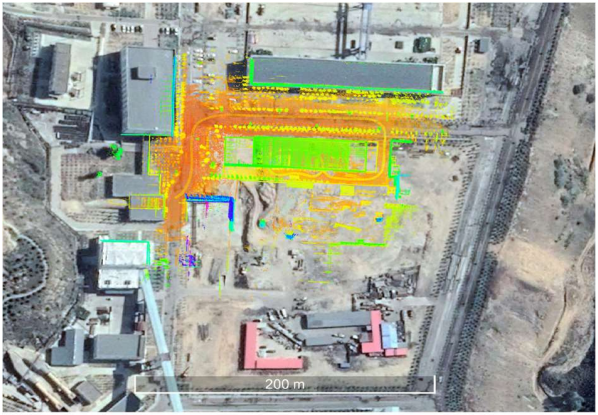

**Fig. 9**. The mapping result projected onto satellite image, the color is coded by height variations.

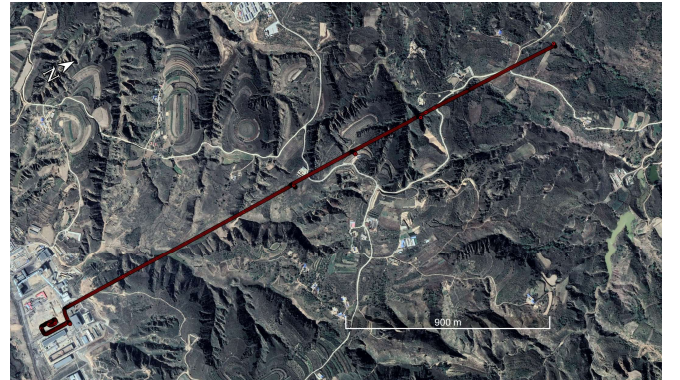

**Fig. 10**. The trajectory of our approach plotted onto satellite image.

*B. Underground Tunnel Tests*

The second experiments seek to evaluate the positioning and mapping accuracy of our system in the mine tunnel. As visualized in Fig. 10, the mine service vehicle stays still outside of the tunnel for RTK and coordinates initialization. Then it enters the tunnel, travels for more than 1700 s, and leaves the tunnel finally, the overall time consumption is 2137 s. The global map is plotted in Fig. 11, in which the top view of the map demonstrate that our map is of high consistency horizontally. Besides, the side view of our map illustrates our result is of high consistency vertically. In addition, we present some of the mapping details both inside and outside of the mine tunnel, the clear and vivid structure on the wall or on the ground demonstrating our mapping is of high precision locally.

The underground tunnel is dominated by flat walls and ground, which is of low texture and repetitive patterns, making it one of the most challenging scenarios for SLAM algorithms. We observe that all the selected methods fail to provide entire state estimation results throughout the tunnel. They either 'stops' at certain areas or fails completely with great errors. On the contrary, our approach can provide seamless and accurate pose estimation result along the path. As pictured in Fig. 12, the GNSS-IMU-odometer tightly coupled output of INS570D can provide continuous pose output regardless of the environment variations. However, the GNSS measurements merely give a one-off correction when satellite signal is available, and the accumulated errors are not corrected where satellite signal is not available. On the other hand, our approach utilizes the detected loop and GNSS positioning to perform the one-off correction. The following accumulated errors are then corrected by ICP-based map matching. In this way, we can correct the errors spread along the trajectory.

To further present the superiority of our system, we plot the absolute error over distance in Fig. 13 for detailed reference. We can directly infer that our method has magnitudes of higher accuracy than that of the INS570D. The reason is two-fold: Firstly, the mine service vehicle has an explosion-proof design, and the tires are made of rubber, which easily lead to wheel slippage in the mine tunnel. Therefore, the wheel odometer is not accurate, especially when the vehicle enters or leaves tunnel branches. Secondly, the accumulated errors are only corrected once when the GNSS signal is available, leaving the remaining errors unsolved. As shown at distance 6600 m in Fig. 13, the errors are corrected with many outliers observable. On the other hand, our loop detection and re-initialization process smooths

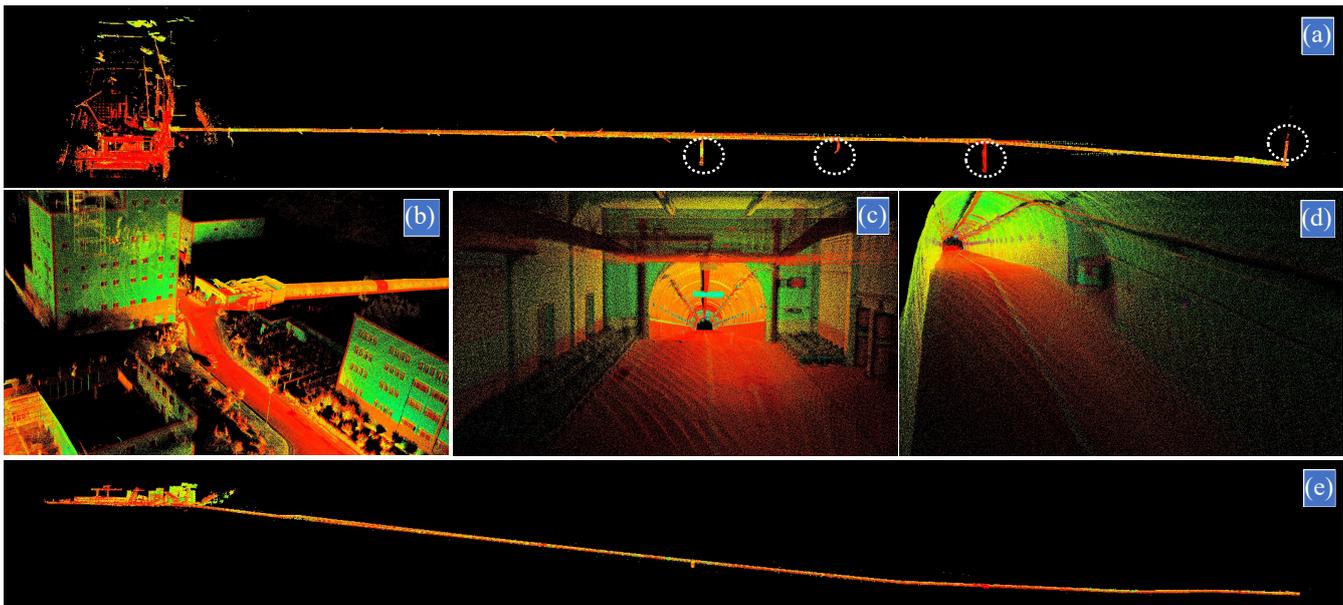

**Fig. 11**. The top view of the global map in (a), and the four dashed white circle indicate the tunnel branches where the vehicle enters and transport staff workers. (b), (c), and (d) are the detailed view of cooperate mapping result, where (b) shows the ground view, (c) presents the entrance of tunnel, (d) gives a tunnel branch. (e) is the side view of the global map.

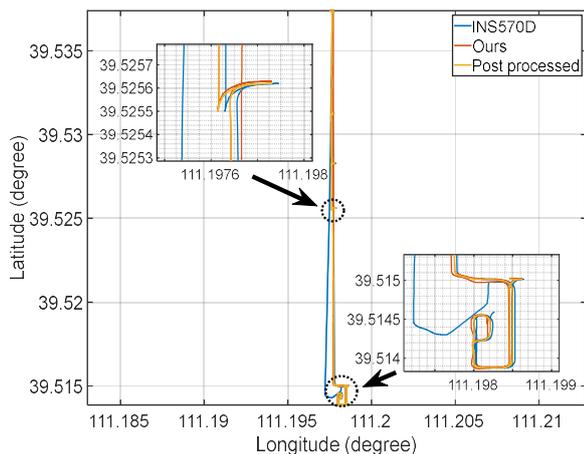

**Fig. 12.** The trajectory comparison of our method and INS570D w.r.t. post processed ground truth.

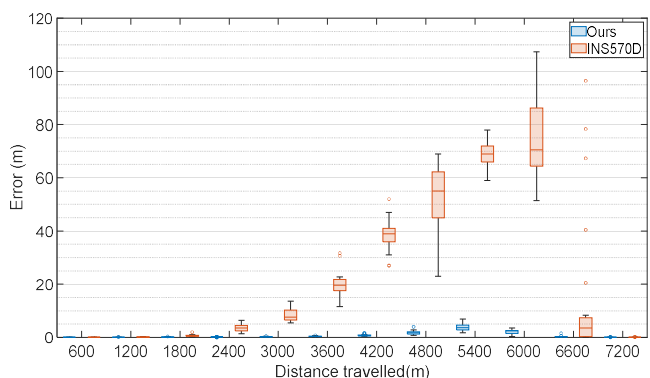

**Fig. 13.** The absolute translational errors over distance.

the whole trajectory bi-directionally. The RMS and MAX error of our system is 2.465 m and 16.861 m.

*C. Consistency Tests*

The third experiments seek to evaluate the consistency of our system within different journeys or datasets. We use GTSAM to perform pose graph optimization, which may generate non-identical pose estimation results even for the same datasets of different trials. The consistency measures the similarity of different paths, which strongly describes the GNSS dropout performance. Besides, the vehicles need to transport staff or necessities to given branch tunnels, the global consistency is of vital importance for such tasks.

We seek to check the consistency within different platforms and trials. In addition to the onboard computer, we use a laptop with Intel i7-10510U CPU, 16 GB RAM for comparison. We perform three runs on each computer for the same dataset, the parameters remain unchanged for different trials. The trajectories of different trials are visualized in Fig. 14, where we can find the paths over various trials does not change much. The maximum trajectory-to-trajectory error is below 20 cm, which is acceptable for cross platform vehicle navigation.

*D. Ablation Study*

In this experiment, we aim to understand the contribution of different sensors and factors. We hereby define the following notations in TABLE II for illustration.

The first test seeks to understand the contribution of different LiDARs. We employ a tunnel-only sequence for illustration, and we plot the individual trajectories in Fig. 15. ours w/o Livox fails shortly in the tunnel and generates large errors. This is due to the flat wall in the tunnel, even the extracted surfel is repetitive, and the point-to-plane pose estimation is not accurate. On the contrary, ours w/o RS survives in this scenario. The reason is twofold: Firstly, the surfel extracted from the surrounding tunnel walls are of high similarity and are harmful for pose estimation. The restricted FoV of Livox Avia is now an advantage in the tunnels, where the observable features are mainly in front of the vehicle and the harmful features are largely omitted. Secondly, Avia has a higher density than RS-16. The increased density leads to more observable features in single scan, where slight environment changes can be detected. Such changes include road signs, holes on the wall, or road curbs in the tunnel, which can provide strong constraint as discussed in our previous work [38]. These two reasons can be interpreted as the DOP in the field of GNSS, where each extracted surfel can be viewed as a satellite. The satellite distribution is now represented using surfel distribution, where Avia has a lower DOP due to better distribution. Therefore, the pose estimation utilizing Avia has an optimal solution. However, the RS-16 is still indispensable of the system. As visualized in the two insets of Fig. 15, when RS-16 is excluded for pose estimation, the system will generate slight errors with irregular motion (sharp turning or forward-backward moving). The omnidirectional view of RS-16 now effectually constrains the pose estimation result, especially for the yaw direction. To further reveal this effect, we plot the two maps around the same tunnel branch for illustration in Fig. 16. It is evident that the global map is consistent and the local map is clear when RS-16 LiDAR is utilized for pose estimation.

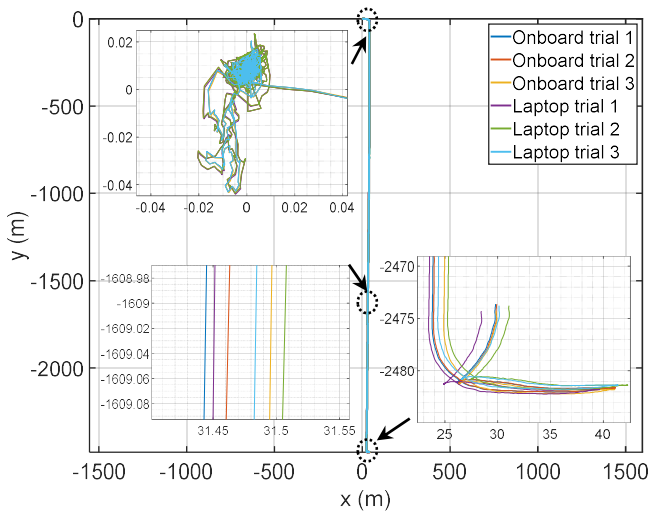

**Fig. 14.** The consistency test across different platforms and trials, and the three insets denote the start, middle, as well as the end of the journey.

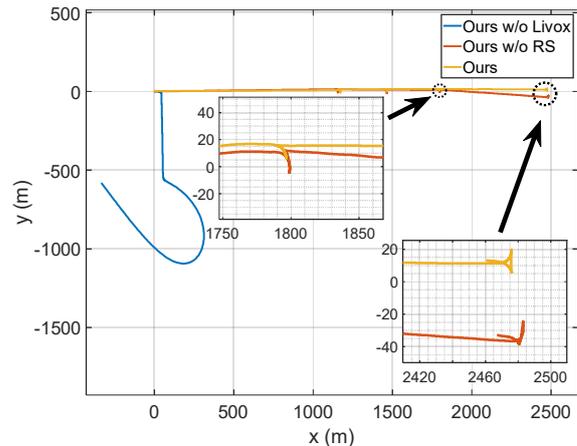

**Fig. 15.** The trajectory comparison of LiDAR contribution ablation study. The two insets denote the errors caused by irregular motions of ours w/o RS.

TABLE II
THE NOTATIONS USED FOR ABLATION STUDY

| Notations | Explanations |
| --- | --- |
| ours w/o Livox | Our SLAM algorithm without Livox Avia LiDAR. |
| ours w/o RS | Our SLAM algorithm without RS-16 LiDAR. |
| ours w/o Surfel | Our SLAM algorithm using feature point matching method for LiDAR odometry. |
| ours w/o RE | Our SLAM algorithm without GNSS reinitialization. |

The second test seeks to reveal the effectiveness of surfel-based scan matching. We also implement a LOAM-like [39] feature points based scan matching approach of our system, denoted as ours w/o Surfel. In addition, we also treat feature points with large intensity variations as edge points as presented in our previous work [18]. Since the surfel based scan matching also suffers from the degeneration for RS-16 as mentioned in the last paragraph, we only select Avia LiDAR for illustration. We select two scenarios, one on the ground and one in the tunnel for comparison. The extracted surfel and feature points in different scenarios are visualized in Fig. 17. It is seen that the edge and planar points have a good distribution on the ground, which can provide strong pose constraint in all direction. However, the planar points have a bad distribution in the tunnel, where only the lateral motion can be constrained, leading to partial unobservability of the longitudinal direction. On the other hand, the extracted surfel have a uniform distribution both on the ground and in the tunnel, which should give better pose estimation results. We hereby utilize a tunnel-only sequence to verify our assumption. As pictured in Fig. 18, the vehicle starts at the middle of tunnel, travels to the last tunnel branch, and returns to the start point, then finally leaves the tunnel. Although ours w/o Surfel maintains consistent and accurate pose estimation for more than 1 km in the tunnel, it still fails to keep this result to the ground due to degeneration problems. On the other hand, the surfel based scan matching can finish the whole journey, and achieves 40% lower return-to-start-point error as visualized in Fig. 18, satisfying with our hypothesis.

The third test seeks to understand the contribution of GNSS re-initialization stage. The sequence used is the same with the underground tunnel test. We denote ours w/o RE as our system merely utilizing GNSS pose graph optimization without re-initialization stage. When re-initialization is not included, our system will directly integrate the pseudorange and carrier-phase measurement into joint state estimation when integer ambiguity is at the fixed solution. Although the maintained pose graph is updated upon the GNSS measurements, the accumulated errors are not eliminated completely as pictured in Fig. 19(a). This is because ours w/o RE uses the same coordinate transformation matrix along the journey, which is approvable when GNSS is always available or temporally unavailable. However, this impact is not negligible in the case of longer dropouts, while the

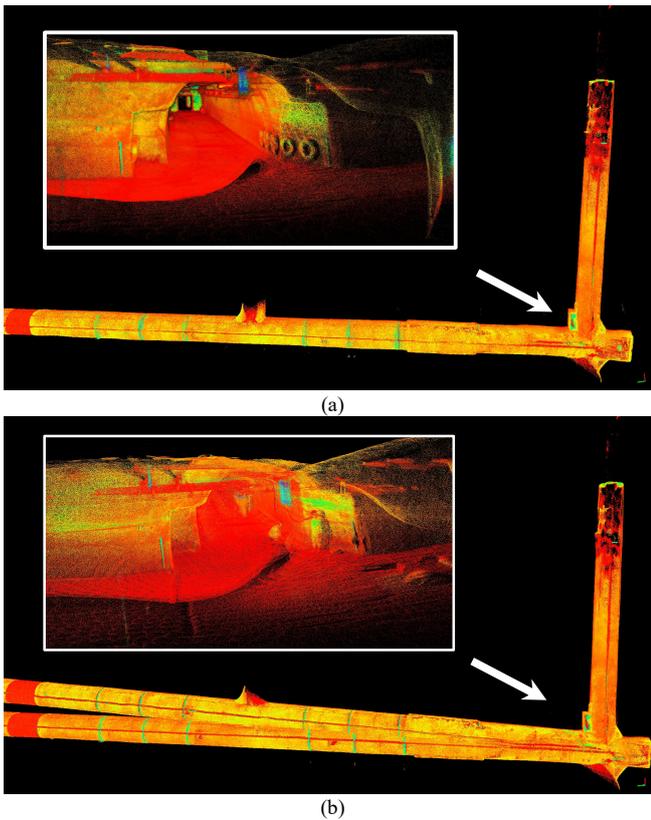

**Figure 16.** Visualization of the mapping difference without Robosense RS-16 LiDAR. (a) is the result of our approach whereas (b) is from ours w/o RS. The two insets in (a) and (b) presents the mapping details of a tunnel branch.

SLAM algorithms suffer from drift during the absence of GNSS measurements. Therefore, there is a large deviation between the pose estimation of SLAM and GNSS, leading to not fully compensated states as visualized in Fig. 19(a). If we manually increase the weight of GNSS measurements, the drift outdoor can then be fully corrected. However, the trajectory is not coherent with several bulges caused by this "forced" optimization as visualized in Fig. 19(b). This non-consistent state also leads to large map distortion, especially in the rolling direction, where the tunnel is rotated for more than 30 degrees.

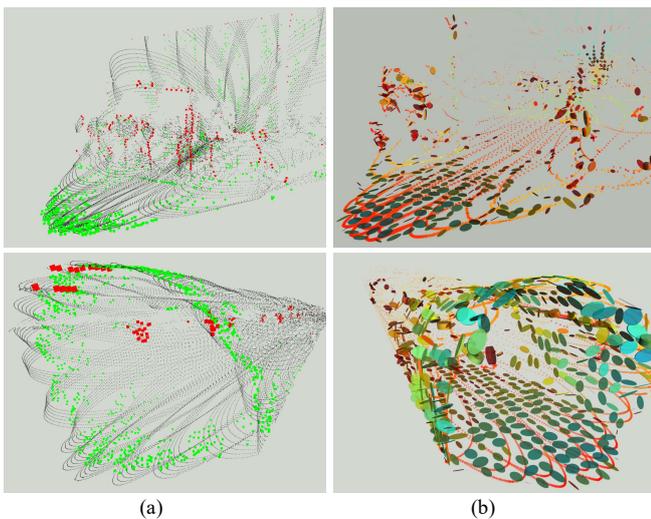

**Fig. 17.** Visualization of the extracted feature points in (a) and surfel in (b), the top two are on the ground while the bottom two are in the tunnel. The red and green points in (a) are the extracted planar and edge points. The markers in (b) are the extracted surfel.

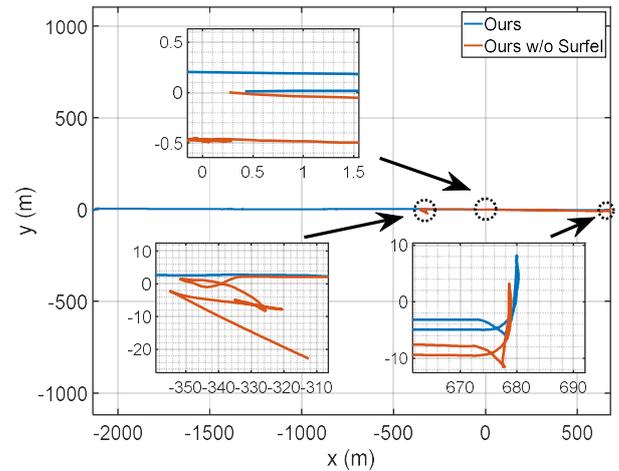

**Fig. 18.** The trajectory comparison of surfel based and feature points based LiDAR SLAM. The three insets denote the starting area, tunnel branch and places where feature points based approach has large outliers.

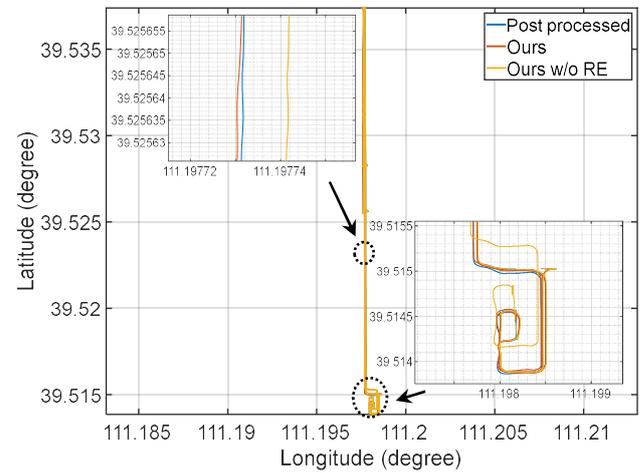

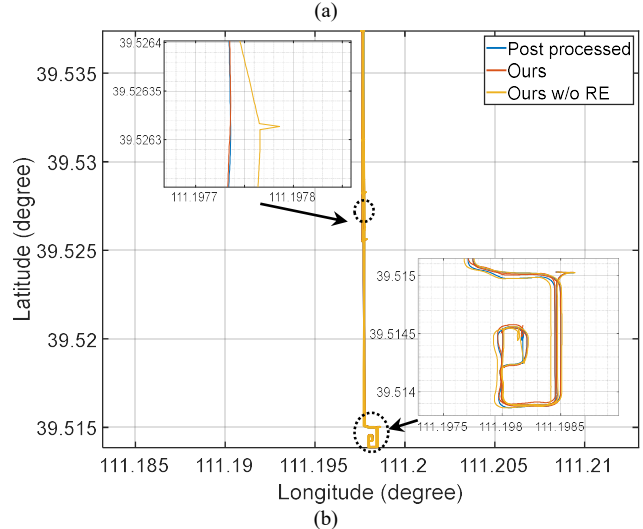

**Fig. 19.** The trajectory comparison of our approach with or without GNSS re-initialization stage. Ours w/o RE has a lower GNSS weight in (a), where drifts outdoor are not fully compensated. On the other hand, the GNSS weight is higher in (b), where many bulges exist on the path. The upper inset in (b) denote an example of such bulges

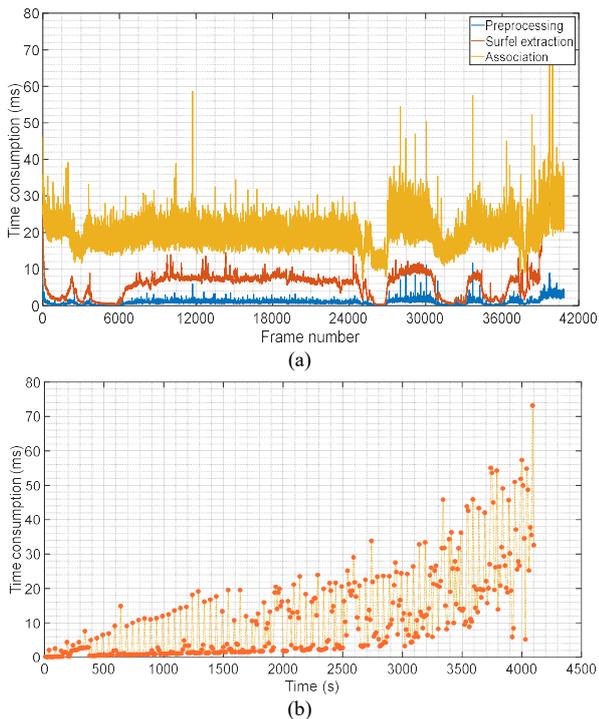

**Fig. 20.** The time consumption of the LiDAR odometry part of our system in (a) and the state optimization part in (b).

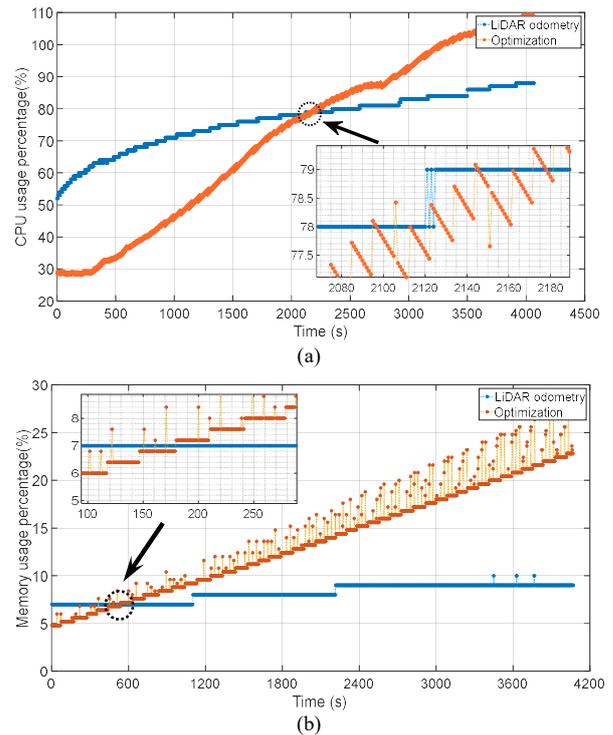

**Fig. 21.** The CPU and memory usage of LiDAR odometry thread and state optimization thread in (a) and (b).

*E. Runtime Efficiency*

Since our system is designed to provide state estimation results to the autonomous mine service vehicles, the runtime efficiency is of prime concern for real-world deployment. We thereby drive the vehicle into the tunnel, travels for 4067 s, and record the per-module time consumption along the journey. Note that our algorithm only utilizes two cores out of the six cores of the ARM Carmel CPU (the internal CPU of Xaiver NX).

We plot three key processes in the LiDAR odometry in Fig. 20(a). The preprocessing stage include outlier and distortion removal, down sampling, and voxelizing. Then the surfel extraction and association denote the time consumption of surfel feature extraction and surfel matching process. It is seen the LiDAR odometry can reach real-time performance even on this embedded ARM CPU, the average runtime is 27.20 ms. Since the real-time map rendering is too time-consuming for this power-limited platform, we disable the visualization thread and plot the state optimization time usage in Fig. 20(b). We can infer that the time used for optimization is increasing continuously along the journey due to enlarged pose graph. Besides, when the vehicle drives out of the tunnel, the loop detection and re-initialization thread is invoked, leading to a large increment after 3500 s. The average time consumption inside the tunnel is 8.12 ms, whereas that of the leaving tunnel is 29.57 ms. Note that the time consumption of IMU/odometer preintegration is negligible, and we do not consider this impact in the experiment.

We also record the CPU and memory usage along the journey. Since our LiDAR odometry and state optimization process are running in different threads, we continuously record the thread statistics as shown in Fig. 21. It is seen that the CPU and memory usage of LiDAR odometry thread is almost steady for the entire journey. Since our memory is 16 GB, the average memory consumption of LiDAR odometry is 1.31 GB. On the other hand, the GTSAM-based graph optimization is constantly occupying computational resources due to increased graph size.

V. CONCLUSION

In this paper, we proposed a localization and mapping framework for autonomous mine service vehicles, achieving accurate and robust pose estimation results in such scenes. Our system integrates measurements from two LiDARs, an IMU, wheel odometers, and optionally a GNSS receiver in a tightly-coupled manner. The front-end includes two parallel running ESKF based LiDAR-inertial odometry. Different from common algorithms utilizing feature points, we extract surfel elements for scan-to-scan registration. Pose results from different estimation engine are jointly optimized at the back-end using pose graph optimization. To fully alleviate the long-term accumulated drift in the tunnel, also known as the GNSS dropouts, we utilize a loop detection based re-initialization process for state alignment. The proposed method has been extensively validated in real-world mine environments, with an acceptable accuracy in most scenarios. In addition, our system has been successfully deployed for several autonomous mine service vehicles for state estimation.

There are several directions for future research. Check-point based mapping evaluation in tunnels is desirable, which helps to understand the system mapping performance. Another research direction concerns developing a safer and easier-to-use platform for large deployment. Many engineering designs, such as explosion-proof shell and long during continuous operation, need to be considered.

ACKNOWLEDGMENT

We would like to thanks the Suzhou plusgo CO., Ltd and Tsinghua University Suzhou Automotive Research Institute for program support and data collection.

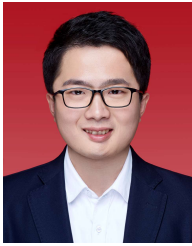

**Yusheng Wang** received the B.Eng. degree in navigation engineering from Wuhan University, Wuhan, China, in 2016, and the M.S. degree in geodesy engineering from the Stuttgart University, Stuttgart, Germany, in 2018. He is currently working towards the Ph.D. degree with the GNSS research center, Wuhan University under the supervision of Prof. Yidong Lou.

He worked as a software engineer in Wuhan In-Driving Co., Ltd from 2018 to 2020, where he was in charge of the localization department. He is a co-founder of Beijing Lishedachuan Co., Ltd and also a senior SLAM engineer in CHCNAV.

His research interests include sensor fusion, SLAM in industrial application, and computer vision.

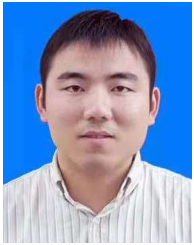

**Yidong Lou** received the B.Eng. degree in geodesy engineering from Wuhan University, Wuhan, China, in 2001, and the M.S. degree in geodesy engineering from Wuhan University in 2004, and the Ph. D degree in geodesy engineering from Wuhan University in 2008.

He is currently with the GNSS research center, Wuhan University, as a professor. His research activity focuses on the theoretical methods and software of GNSS real-time high-precision data processing as well as the meteorological applications. His research findings have been successfully applied to the major project, "National Beidou Ground-based Augmentation System Development and Construction", which has realized the wide-area real-time positioning accuracy of Beidou from meter level to centimeter level, supporting the innovative applications of a nationwide network service.

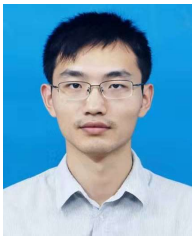

**Weiwei Song** received the B.Eng., M.S., and Ph.D. degrees in geodesy engineering from Wuhan University, Wuhan, China, in 2004, 2007, and 2011, respectively. He is currently with the GNSS Research Center, Wuhan University, as a professor. His research interests include sensor fusion in industrial application and high-precision GNSS localization.

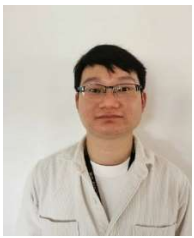

**Bing Zhan** received the B.Eng. degree in navigation engineering from Wuhan University, Wuhan, China, in 2016. Then he worked as a product manager in CHCNAV. He is now a senior product manager of monitoring department in CHCNAV.

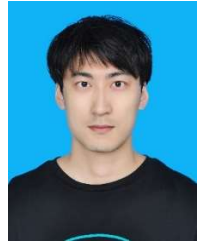

**Feihuang Xia** received the B.Sc. from Zhejiang University, Zhejiang, China, in 2017 and Ph.D. degree in Peking University in 2021. He is currently with the Beijing Lishedachuan CO., Ltd. His research interest is manifold math and SLAM.

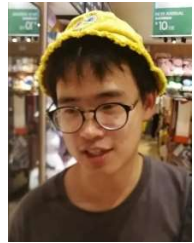

**Qigen Duan** received the B.Eng. from China University of Geosciences, Beijing, China, in 2017. He is currently working towards the Ph.D. degree with the Department of Geography and Resource Management, Chinese University of Hongkong, under the supervision of Prof. Hui Lin.

He worked as a software engineer in NavInfo Co., Ltd from 2017 to 2018. He is also the founder of Beijing Lishedachuan Co., Ltd.

His research interests include SLAM, 3D reconstruction, and digital twin.